\documentclass[letterpaper,10pt,journal,twoside]{IEEEtran}
\usepackage{amsmath,amsfonts}
\usepackage{algorithmic}
\usepackage{algorithm}
\usepackage{array}
\usepackage[caption=false,font=normalsize,labelfont=sf,textfont=sf]{subfig}
\usepackage{textcomp}
\usepackage{stfloats}
\usepackage{url}
\usepackage{verbatim}
\usepackage{graphicx}
\usepackage{cite}
\usepackage{xcolor}
\usepackage[shortlabels]{enumitem}
\usepackage[hidelinks]{hyperref}
\usepackage{amsfonts}
\usepackage{amsmath}
\usepackage{amsthm}
\usepackage{amssymb}
\newcommand{\real}{\mathbb{R}}
\newcommand{\wrench}{\mathcal{F}}
\newcommand{\twist}{\mathcal{V}}

\begin{document}

\title{Human-Multirobot Collaborative Mobile Manipulation: the Omnid Mocobots}
\author{Matthew L. Elwin,~\IEEEmembership{Member}, Billie Strong,~\IEEEmembership{Member}, Randy A. Freeman,~\IEEEmembership{Member}, and Kevin M. Lynch\footnote{blah},~\IEEEmembership{Fellow}% <-this % stops a space
\thanks{This work was supported by Northwestern University.}
\thanks{The authors are with the Center for Robotics and Biosystems, Northwestern University, USA: 
  {\tt\footnotesize elwin@northwestern.edu}}
}%

\maketitle

\begin{abstract}
The Omnid human-collaborative mobile manipulators are an experimental platform for testing control architectures for autonomous and human-collaborative multirobot mobile manipulation. An Omnid consists of a mecanum-wheel omnidirectional mobile base and a series-elastic Delta-type parallel manipulator, and it is a specific implementation of a broader class of mobile collaborative robots (``mocobots'') suitable for safe human co-manipulation of delicate, flexible, and articulated payloads. Key features of mocobots include passive compliance, for the safety of the human and the payload, and high-fidelity end-effector force control independent of the potentially imprecise motions of the mobile base. We describe general considerations for the design of teams of mocobots; the design of the Omnids in light of these considerations; manipulator and mobile base controllers to achieve multirobot collaborative behaviors; and experiments in human-multirobot collaborative mobile manipulation of large and articulated payloads, where the mocobot team renders the payloads weightless for effortless human co-manipulation. In these experiments, the only communication among the humans and Omnids is mechanical, through the payload.
\end{abstract}


\section{Introduction}
\label{sec:intro}

\begin{figure}
    \centering
    \includegraphics[width=3.4in]{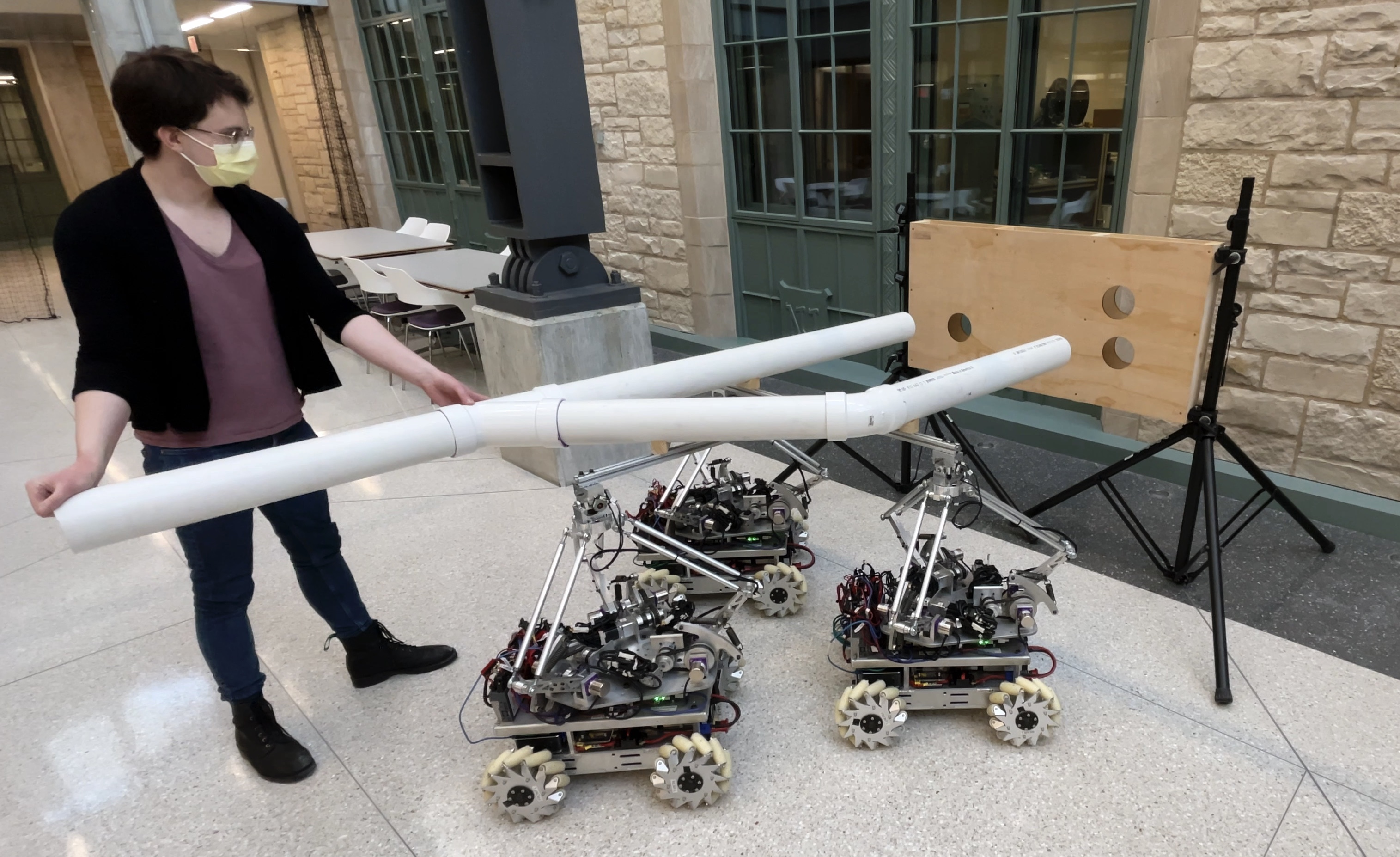}
    \caption{A human and three Omnid mocobots performing 6-dof collaborative assembly of a bulky payload into a fixture.}
    \label{fig:billie}
%    \vspace{-10pt}
\end{figure}

This paper introduces the Omnid mobile collaborative robot, or ``mocobot'' for short. The Omnid mocobots are designed specifically for team mobile manipulation, including autonomous cooperative manipulation and manipulation in collaboration with one or more human partners (Figure~\ref{fig:billie}).

Here we focus on human-multirobot collaborative mobile manipulation, particularly factory, warehouse, or construction manipulation tasks with the following characteristics: 
\begin{enumerate}[(a)]
    \item the payload is large and it is impractical for a single robot to manipulate it;
    \item it is advantageous to distribute contact forces over the payload, perhaps to minimize stress concentrations on a delicate payload or to control all the degrees of freedom of an articulated or flexible payload; and
    \item the task is unique or difficult to automate, requiring the adaptability and situational awareness of one or more human collaborators.
\end{enumerate}
Even if tasks are eventually automated, a human-collaboration phase can be useful for human-guided machine learning.

Human-robot communication during collaboration can be verbal, visual, or physical, but here we focus on physical human-robot interaction through the payload because the direct haptic interaction provides the human a high-bandwidth and intuitive communication channel for manipulation tasks~\cite{Kucuktabak2021}. 
The mocobots also communicate with each other mechanically through the payload. 
Although other types of communication among the robots and between the humans and robots are possible, and can facilitate other capabilities, the examples in this paper use only physical communication through the payload.

\emph{Paper overview:} 
For the safety of the human and the payload, mocobots should provide passive mechanical compliance to compensate for the limited bandwidth of active impedance and force controllers. Mocobots should also provide sufficient manipulability of the payload to enable effective autonomous or human-collaborative manipulation. We describe these and other design requirements for a team of mocobots (Section~\ref{sec:considerations}); explain how the design of the Omnids, with their omnidirectional mobile bases and series-elastic parallel manipulators, satisfies these requirements (Section~\ref{sec:design}); present controllers that facilitate useful human-collaborative behaviors (Section~\ref{sec:control}); and describe experiments where the Omnids render large payloads weightless for easy collaborative manipulation by one or more human partners (Section~\ref{sec:experiments}). 

\emph{Contribution:} 
The primary contribution of this paper is the development of a novel multi-mocobot system with capabilities that have not been demonstrated by any previous collaborative mobile manipulator team. In particular, we introduce the design and control of the Omnid mocobots that enable one or more humans to physically co-manipulate a payload with all of the following features: 
\begin{enumerate}[(1)]
\item payload manipulation occurs in three-dimensional space;
\item the co-manipulation can be performed by any number of mocobots beyond a minimum number (three for the Omnids) to (i) distribute the forces on the payload to avoid stress concentrations, (ii) achieve sufficiently large forces to support the payload, and/or (iii) control payload internal degrees of freedom; 
\item the mocobots provide high-fidelity active force control as well as passive compliance, which enhances the safety of the humans and the safety of fragile payloads to disturbances and transient control errors; and
\item the mocobot team can cancel the gravitational force on the payload, implementing realistic weightless dynamics.
\end{enumerate}
To our knowledge, no previous work has combined these features in a single system (see Section~\ref{sec:related}). The combination is powerful, as {initial experiments indicate} it allows safe, effortless, dynamic, and intuitive manipulation of large, awkward, and articulated payloads by one or multiple human users. These features are demonstrated in the video accompanying the paper, \url{https://tinyurl.com/omnids}~\cite{OmnidVideo}.

Other contributions include the design of a novel SEA-actuated force-controlled parallel manipulator and the definition of payload manipulability for a multirobot system.

\section{Related Work}
\label{sec:related}

The ``cobot'' concept for human-robot collaborative manipulation was introduced in~\cite{Colgate1996} and focused on ensuring operator safety through the use of programmable constraints. 
Optimal constraints for ergonomic human-cobot interaction, including collaborative mobile manipulation, are described in~\cite{Lynch2002,Pan2005}.

Related previous work includes physical human collaboration with a single mobile manipulator, autonomous manipulation by a team of mobile manipulators, and human collaborative manipulation with a team of mobile manipulators.

\emph{Human collaborative manipulation with a single mobile manipulator:} 
Kosuge et al.~\cite{Kosuge2000} describe the development of an early dual-arm mobile manipulator intended for human interaction called ``MR~Helper,'' where each arm is a 7-DOF Mitsubishi PA-10 equipped with a six-axis wrist force-torque sensor. Many other single- and dual-arm mobile manipulators for human interaction have been developed, including bipeds participating in shared carrying of objects (e.g.,~\cite{Agravante2019,Agravante2014,Bussy2012,Yokoyama2003,giammarino2022superman}).
Approaches to human-robot physical collaboration include active compliance control~\cite{StucklerBehnke2011}, impedance control~\cite{DeCarli2009}, and the identification of a discrete set of load-sharing policies between the human and the robot~\cite{LawitzkyMortlHirche2010}.  Some benefit has been observed for allowing humans and robots to dynamically exchange leader and follower roles during co-manipulation~\cite{Mortl2012,Evrard2009}.  
Machine learning has also been applied to infer human intent during cooperative manipulation~\cite{BergerEtAl2013,BergerEtAl2014,Lanini2018}.

\emph{Autonomous manipulation by a team of mobile manipulators:} Several research groups have demonstrated autonomous transport of a payload using a team of mobile manipulators (e.g.,~\cite{Hirche2020,carey2021collective,WangSchwager2016a,rauniyar2021mewbots,Alonso-Mora2015local}). The mechanical overconstraint of the closed loops is typically accounted for by active impedance control at the manipulators and mechanical compliance of the payload.

\emph{Human collaborative manipulation with a team of manipulators:} There has been little prior work on human-multirobot mobile manipulation. Notable exceptions include systems designed for autonomous multirobot manipulation but modified to replace the robot leader with a human leader. 
In~\cite{Sieber2015}, a human leads a multirobot system manipulating an object via formation control, directing the motion remotely with the assistance of vibrotactile feedback.  In~\cite{WangSchwager2016}, the human teleoperates the leader robot or physically grabs the co-manipulated object, and the leader's force is amplified by the other robots in the team. The motion is restricted to a horizontal plane. In~\cite{sirintuna2022carrying}, an object is carried by multiple mobile manipulators and a human, who partially bears the weight of the object. 

\emph{Contribution of this paper:} Unlike most mobile manipulators that employ commercial robot arms~\cite{Hirche2020,carey2021collective,Alonso-Mora2015local,sirintuna2022carrying}, our Omnid mocobot team is the first designed from the ground up for human-multimocobot collaborative manipulation, including significant mechanical compliance for human and payload safety and high-fidelity active force control. As mentioned in Section~\ref{sec:intro}, the system introduced in this paper is unique in its ability to render payloads weightless, allowing effortless, dynamic, and intuitive manipulation of large and articulated payloads.

\section{Design Considerations for a Mocobot Team}
\label{sec:considerations}

This section describes general design considerations for any mocobot team, while Section~\ref{sec:design} describes how the Omnid mocobots address these considerations.

Many existing mobile manipulators consist of a wheeled mobile base with one or more robot arms (e.g., a UR16~\cite{giammarino2022superman} or a Franka Emika Panda~\cite{Hirche2020}) mounted to it. When multiple mobile manipulators grasp a fragile payload, however, the safety of the load may be at risk due to mechanical overconstraint. The closed-loop kinematic constraints can be modeled by $m$ holonomic constraints $f(q) = 0$, where $q \in \real^n$ is a set of generalized coordinates describing the configuration of the payload and the mobile manipulators. Wheeled robots may also be subject to $k$ nonholonomic constraints of the form $H(q)\dot{q} = 0$. The holonomic constraints can be differentiated and combined with the nonholonomic constraints to obtain a complete set of $m+k$ Pfaffian constraints of the form $A(q)\dot{q} = 0$, and the dynamics of the multirobot-payload system can be written collectively in the form
\begin{align}
    M(q)\ddot{q} + c(q,\dot{q}) + p(q) & = S(q)u(q) + A^\intercal(q) \lambda \label{eq:stdeqs} \\
    A(q) \dot{q} & = 0, \label{eq:Pfaffian}
\end{align}
where $M(q) \in \real^{n \times n}$ is the combined mass matrix of the robots and the payload; $c(q,\dot{q}) \in \real^n$ is a vector of Coriolis, centripetal, and possibly mechanical damping generalized forces; $p(q) \in \real^n$ is a vector of generalized forces due to potential (e.g., gravity or springs); $u(q) \in \real^{n_{\textrm{a}}}$ is the $n_{\textrm{a}}$-vector of robot control forces and torques produced by a feedback controller (which is typically at least a function of $q$); $S(q) \in \real^{n \times n_{\textrm{a}}}$ is a transmission matrix mapping the controls to the generalized coordinates they act on; $A(q) \in \real^{(m+k) \times n}$ represents the Pfaffian constraints; and $\lambda \in \real^{m+k}$ is a set of generalized forces enforcing the constraints. See~\cite{Hirche2020} for an example of such a model for multirobot manipulation.

One goal of the feedback law $u(q)$ could be to minimize components of the constraint forces $\lambda$ associated with the grasps, avoiding large compressive or tensile internal forces on the payload. The simplified dynamics~\eqref{eq:stdeqs} do not distinguish, however, between passive mechanical forces (e.g., $p(q)$) and the active feedback-controlled forces $u(q)$, which are subject to limited bandwidth and possibly network latency. The non-idealities of active feedback control mean that incompatibilities of the motions of the end-effectors of the mocobots must be absorbed by mechanical compliances in the robots or the payload itself. These mechanical compliances are often poorly understood, and a fragile payload may be damaged.

\subsection{Passive Compliance}
\label{ssec:passivecompliance}

For the safety of the payload and human collaborators, mocobots should implement well-characterized passive mechanical compliance. This compliance may be achieved by passive elements (e.g., springs) or actuators such as direct-drive actuators, series-elastic actuators (SEAs), variable stiffness actuators (VSAs), or variable impedance actuators (VIAs). With mechanical compliance, the end-effector of the $i$th mocobot exhibits a passive stiffness that can be locally linearized as
\[
\delta \wrench_i = K_i \delta X_i,
\]
where the (possibly configuration-dependent) positive-semidefinite stiffness matrix $K_i \in \real^{6 \times 6}$ maps  $\delta X_i \in \real^6$ (a differential change in the exponential coordinates describing the configuration of the end-effector) to $\delta \wrench_i \in \real^6$ (a differential change in wrench), where $X_i$ and $\wrench_i$ are expressed in a common end-effector frame. If $N$ mocobots grasp a rigid payload, the total stiffness of the payload can be expressed in a common payload frame as
\begin{equation}
K = \sum_{i=1}^{N} [{\rm Ad}_{T_{i\textrm{p}}}]^\intercal K_i [{\rm Ad}_{T_{i \textrm{p}}}],
\label{eq:stifftrans}
\end{equation}
where $T_{i \textrm{p}} \in SE(3)$ expresses the configuration of the payload relative to the $i$th end-effector frame and $[{\rm Ad}_{T_{i\textrm{p}}}]$ is the adjoint representation of $T_{ip}$~\cite{ModernRobotics}.

A necessary condition for the safety of a payload to position perturbations at each end-effector is an upper bound on each end-effector's passive stiffness $K_i$ (e.g., on the maximum eigenvalue of $K_i$), and a necessary condition for the safety of a human interacting with the payload is an upper bound on the total stiffness $K$ of the payload.
Lower bounds on each end-effector's stiffness may come from 
%physical limits on the actuators (e.g., the need to keep springs within their valid length range) and 
bandwidth requirements of some control modes (e.g., motion control).

\subsection{Payload Manipulability}
\label{ssec:manipulability}

Manipulability analysis can be used to understand which payload degrees-of-freedom a mocobot team can control.
Solving Equation~\eqref{eq:stdeqs} for $\ddot{q}$, substituting into a differentiated version of Equation~\eqref{eq:Pfaffian}, solving for the constraint forces $\lambda$, and eliminating $\lambda$ from Equation~\eqref{eq:stdeqs} yields
\begin{equation}
    \ddot{q} = F(q) u + g(q,\dot{q}), 
\end{equation}
where, after dropping the dependence on $q$ for brevity,
\begin{equation}
    F = M^{-1}(I - A^\intercal(AM^{-1}A^\intercal)^{-1}AM^{-1}) S. \label{eq:Fmatrix}
\end{equation}

The full configuration $q \in \real^n$ can be decomposed as $q = [q_{\textrm{p}}^\intercal \; q_{\textrm{r}}^\intercal]^\intercal$, where $q_{\textrm{p}} \in \real^{n_{\textrm{p}}}$ and $q_{\textrm{r}} \in \real^{n_{\textrm{r}}}$ are generalized coordinates of the payload and robots, respectively, and $n = n_{\textrm{p}} + n_{\textrm{r}}$. If the payload is a rigid body, then $n_{\textrm{p}} = 6$, and if the payload is articulated with $d$ internal degrees of freedom, then $n_{\textrm{p}} = 6+d$.

Accordingly, the matrix $F(q)$ from Equation~\eqref{eq:Fmatrix} becomes
\[
F(q) = \begin{bmatrix}
F_{\textrm{p}}(q) \\
F_{\textrm{r}}(q)
\end{bmatrix},
\]
where $F_{\textrm{p}}(q) \in \real^{n_{\textrm{p}} \times n_{\textrm{a}}}$ and $F_{\textrm{r}}(q) \in \real^{n_{\textrm{r}} \times n_{\textrm{a}}}$. This matrix relating controls $u$ to accelerations $\ddot{q}$ depends on the configuration of the payload $q_{\textrm{p}}$ and the robots' configuration $q_{\textrm{r}}$, which determines their grasp locations on the payload. We define the linear \emph{manipulability} of the payload at $q$ as $\operatorname{rank}(F_{\textrm{p}}(q))$, the number of degrees of freedom of the payload that can be instantaneously independently controlled. We say that the payload is fully manipulable at $q$ if $\operatorname{rank}(F_{\textrm{p}}(q)) = n_{\textrm{p}}$.

\subsection{Manipulation Force Control Independent of Mobile Base Motion Control}
\label{ssec:decoupled}

The mobile base of a mocobot ideally keeps its manipulator near the center of its workspace throughout the manipulation, for maximum control authority. We assume the mobile base is motion controlled, perhaps imprecisely. For example, wheels may slip or skid; the footholds of a legged mobile base may be uncertain or yielding; and the ground may be uneven.

Conversely, we require the mocobot to be capable of predictable compliance and force control at its end-effector, effectively independent of the motion of the mobile base, for the safety of the human and the payload. 

Thus the mechanical design and control of the mocobot should support reasonable decoupling between  the force-controlled behavior visible at the end-effector and the potentially imprecise motion control of the mobile base. With this abstraction enforced by design and/or low-level control behaviors, mocobots of different designs (wheeled, tracked, legged, etc.) may implement the same high-level coordinated behaviors for human-collaborative and autonomous team manipulation. Below we describe how the Omnid mocobot effectively implements force control at its end-effector. 

\section{Design of the Omnid Mocobots}
\label{sec:design}
% Introduce the omnid as a modular system comprising a mobile base, delta manipulator, and passive gimbal.

The Omnid mocobots are designed to approximately decouple the force-controlled behavior at the end-effector from imprecise motion control of the mobile base (Section~\ref{ssec:decoupled}), exhibit well-characterized passive compliance (Section~\ref{ssec:passivecompliance}), and, with a team of three or more, achieve full manipulability of payloads with up to three internal degrees of freedom (Section~\ref{ssec:manipulability}). The Omnids are specifically designed for team manipulation; a single Omnid is incapable of controlling all six degrees of freedom of a rigid payload.

\begin{figure}
  \begin{center}
 \includegraphics[width=3.1in]{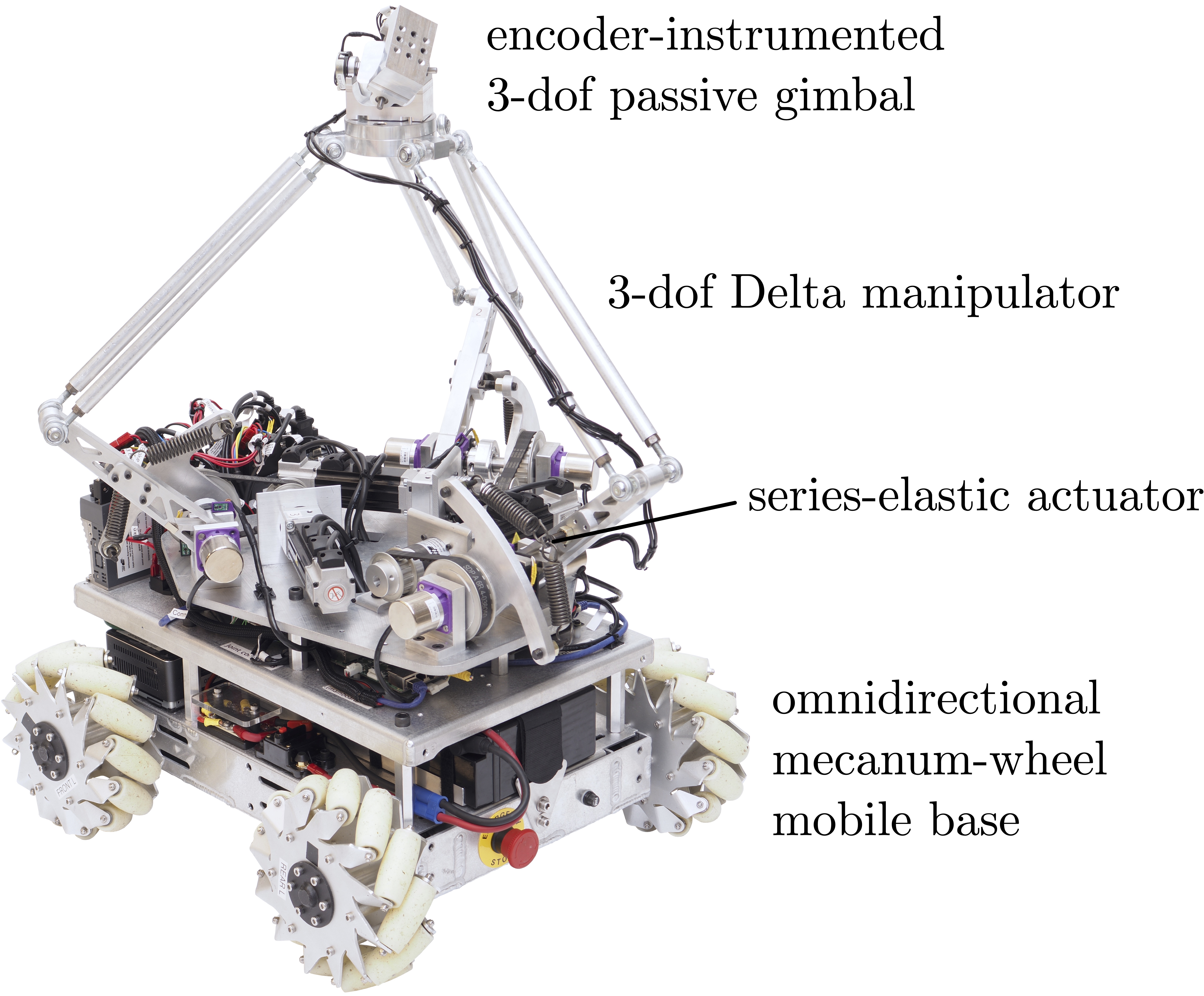}
\end{center}
 \caption{An Omnid mocobot, consisting of an omnidirectional mobile base, a 3-dof SEA-driven Delta parallel manipulator, and a 3-dof passive gimbal.}
 \label{fig:omnid}
\end{figure}

Each Omnid consists of a mecanum-wheel omnidirectional mobile base and a 6-dof manipulator. The manipulator is a 3-dof Delta parallel mechanism~\cite{ModernRobotics} driven by SEAs plus an encoder-instrumented 3-dof passive gimbal ``wrist'' (Figure~\ref{fig:omnid}). Thus an Omnid has six independently-controlled degrees of freedom (two translational and one rotational freedom for the mobile base, plus three translational freedoms for the manipulator) and three passive rotational freedoms at the wrist.

\begin{table}
\begin{center}
\begin{tabular}{|c|c|c|}
\hline
& & stated manipulability is valid \\
     \# Omnids & payload manipulability & if the end-effector(s) are:  \\ \hline
     1  & $\textrm{rank}(F_{\textrm{p}}(q)) = 3$ & anywhere \\
     2  & $\textrm{rank}(F_{\textrm{p}}(q)) = 5$ & not collocated\\
     3 & $6 \leq \textrm{rank}(F_{\textrm{p}}(q)) \leq 9$ & not collinear  \\
     \hline
\end{tabular}
\end{center}
\caption{Manipulability of the payload as a function of the number of Omnids grasping the payload and the configurations of their end-effectors.}
\label{table:manipulability}
%\vspace{-15pt}
\end{table}

With this design, each Omnid can apply three-dimensional linear forces that act on the payload at the center of each wrist. Assuming the end-effectors of Omnids collaboratively manipulating a payload are not at the boundaries of their workspaces, the manipulability of the payload as a function of the number of Omnids grasping the payload is given in Table~\ref{table:manipulability}.  If the gimbal centers of three collaborating Omnids are not collinear, they can locally actuate all six degrees of freedom of a rigid payload, plus up to a maximum of three additional internal degrees of freedom of an articulated payload. If the gimbals of the Omnids are collinear, they cannot resist torque about the line. At such a singularity, the manipulability of a rigid payload drops to $5$.

\subsection{Series-Elastic Delta Manipulator with Gimbal Wrist}

\begin{figure}
    \centering
    \includegraphics[width=3in]{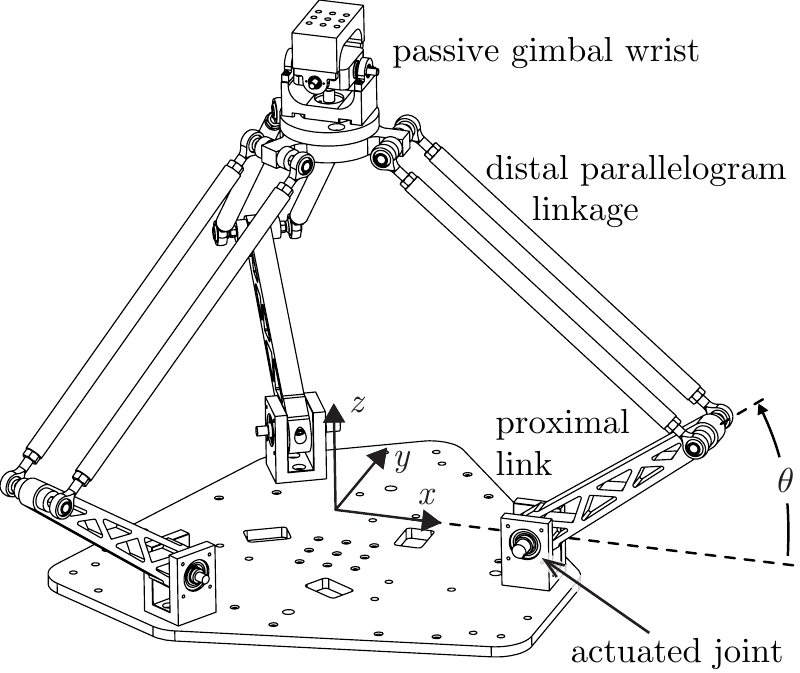}
    \caption{The Omnid's manipulator is an SEA-driven Delta parallel mechanism plus a passive gimbal wrist. The gimbal center is at $\textrm{x} = (x,y,z)$ in the manipulator frame, which is centered between the actuated (proximal) joints.}
    \label{fig:delta}
\end{figure}

The Omnid manipulator is a Delta parallel manipulator plus a passive gimbal wrist (Figure~\ref{fig:delta}). Each of the three legs supporting the Delta's end-effector consists of a proximal revolute joint (driven by an SEA, described below) and a distal unactuated parallelogram linkage. Away from singularities of the mechanism, in the normal operating workspace of the wrist, the angles of the three proximal revolute joints $\theta=(\theta_1,\theta_2,\theta_3)$ map to the three translational coordinates $\textrm{x} = (x,y,z)$ of the center of the gimbal wrist, expressed in a frame fixed to the manipulator's base, through the forward kinematics $\textrm{x} = h(\theta)$. The Jacobian $J(\theta) = \partial h /\partial \theta \in \real^{3\times 3}$ is full rank in this region and satisfies $\dot{\textrm{x}} = J(\theta) \dot{\theta}$ and $\dot{\theta} = J^{-1}(\theta) \dot{\textrm{x}}$. The physical parameters of the Omnid SEA Delta manipulator are given in Table~\ref{tab:params}.

\begin{figure}
  \begin{center}
 \includegraphics[width=3in]{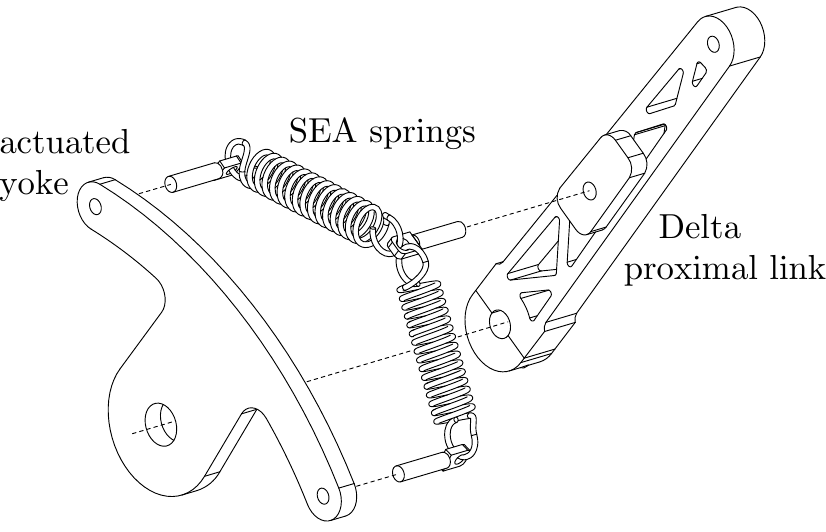}
\end{center}
 \caption{The custom SEA at each proximal joint of the Delta manipulator consists of a yoke driven by a geared motor and two antagonistic linear extension springs.}
 \label{fig:sea-diagram}
\end{figure}

\begin{figure}
    \centering
\includegraphics[width=2.5in]{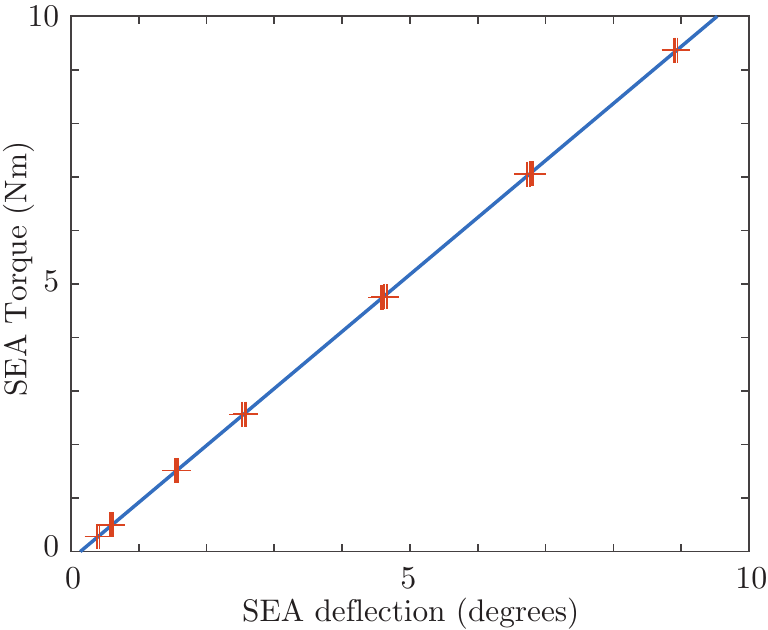}
\caption{Experimental SEA torque as a function of deflection (actuator angle minus the proximal joint angle) and a near-perfect linear fit to the data. }
    \label{fig:stiffness}
\end{figure}

\begin{table}
\begin{center}
\begin{tabular}{|rl|}
\hline
    SEA max continuous torque & $9$~Nm \\
    SEA stiffness & $60.1$~Nm/rad \\ % \hline
    Delta proximal link length &  $0.200$~m \\ % \hline
    Delta distal parallelogram linkage length  &  $0.368$~m \\ % \hline
    Joint limits at each proximal joint $\theta_i$ & $[-15^\circ,100^\circ]$ \\ % \hline
    Joint angles $\theta_i$ at home & $36.6^\circ$ \\ % \hline
    End-effector $(x,y,z)$ position at home & $(0,0,0.420~\textrm{m})$  \\ % \hline
    Radius of workspace-inscribed sphere about home & $0.15$~m \\
    End-effector max continuous force at home & $90$~N \\
    End-effector linear stiffness $K_{xx}, K_{yy}$ at home & 1400 N/m \\ % calculated as 1396 N/m
    End-effector linear stiffness $K_{zz}$ at home & 2000 N/m \\ % calculated as 1985 N/m
    Theoretical end-effector position resolution at home & $0.3$~$\mu$m \\
    Theoretical end-effector force resolution at home & $500$~$\mu$N \\ \hline
\end{tabular}
\end{center}
\caption{Physical properties of the Omnid Delta manipulator and its theoretical position and force capabilities at the home configuration. 
}
\label{tab:params}
%\vspace{-15pt}
\end{table}
Each SEA consists of a brake-equipped Applied Motion J0100-353-3-000 motor with a 40PE025 25:1 planetary gearhead, followed by a toothed 12:7 belt reduction driving a revolute joint coupled to the Delta's proximal revolute joint by two antagonistic linear extension springs (Figure~\ref{fig:sea-diagram}). This setup creates a joint with a maximum continuous torque of approximately $9$~Nm and a highly linear torsional stiffness, approximately $k = 60.1$~Nm/rad (Figure~\ref{fig:stiffness}). Defining the matrix $K_\theta = \operatorname{diag}(k,k,k) \in \real^{3 \times 3}$, the configuration-dependent linear stiffness at the gimbal wrist is $K(\theta) = J^{-\intercal}(\theta) K_\theta J^{-1}(\theta)$. The maximum continuous end-effector force and the stiffness at the home configuration are given in Table~\ref{tab:params}.

The angles of the driving joint and the Delta proximal joint (before and after the springs, respectively) are measured by Broadcom AS38-H39E-S13S 23-bit absolute encoders. Joint torque is estimated by multiplying the joint stiffness by the angular difference between these joints, resulting in a theoretical resolution of approximately 45~$\mu$Nm. See Table~\ref{tab:params} for the resulting end-effector position and force resolution at the home configuration. 

The accurate force sensing and control of the SEA Delta manipulator achieves the desired decoupling of end-effector forces from the motion of the mobile base (Section~\ref{ssec:decoupled}).

The freely-rotating gimbal wrist prevents transmission of torques about the gimbal center. The three angles of the gimbal $\alpha = (\alpha_x,\alpha_y,\alpha_z)$ are measured by CUI AMT213A-V capacitive 12-bit absolute encoders. If the end-effector is attached to a rigid payload, and the location of the end-effector frame relative to the payload frame is known, then the configuration of the payload relative to the Omnid's mobile base is known from the six encoders of the Delta and gimbal. 

\subsection{Omnidirectional Mobile Base}
The omnidirectional mobile base is a modified SuperDroid IG52-DB4 platform driven by four mecanum wheels. Each wheel is driven in velocity-control mode by an encoder-instrumented motor with a gearbox and belt drive. The encoders enable odometry which approximately tracks the incremental motion of the mobile base. For global positioning, an overhead camera can track the payload and/or mobile base, or the mobile base may be equipped with Intel T265 cameras. Global positioning is not needed for many collaborative manipulation modes, however. 

\subsection{Computing and Communication}
Each Omnid is equipped with several TM4C123GH6PM microcontrollers. 
The Delta manipulator has a top-level microcontroller to perform kinematics and whole-arm control at 100~Hz. Each SEA joint has its own microcontroller to implement low-level torque control at 800~Hz, and the gimbal has a microcontroller to measure the gimbal joint angles.  The top-level Delta microcontroller communicates with the four other microcontrollers via dedicated RS-485 buses. 

The mobile base uses three microcontrollers: one for the mobile base controller and one each for low-level control of the front and rear wheels, respectively. The top-level controller communicates with the wheel controllers via RS-485.   

The top-level Delta and mobile base microcontrollers communicate via RS-485 with an Intel NUC7i7BNH PC running Linux and ROS Noetic~\cite{Quigley09}, which handles Wi-Fi communication among robots and a base station. Control modes described in this paper do not require wireless communication.

\section{Omnid Behaviors and Control}
\label{sec:control}
\begin{figure}
  \begin{center}
\includegraphics{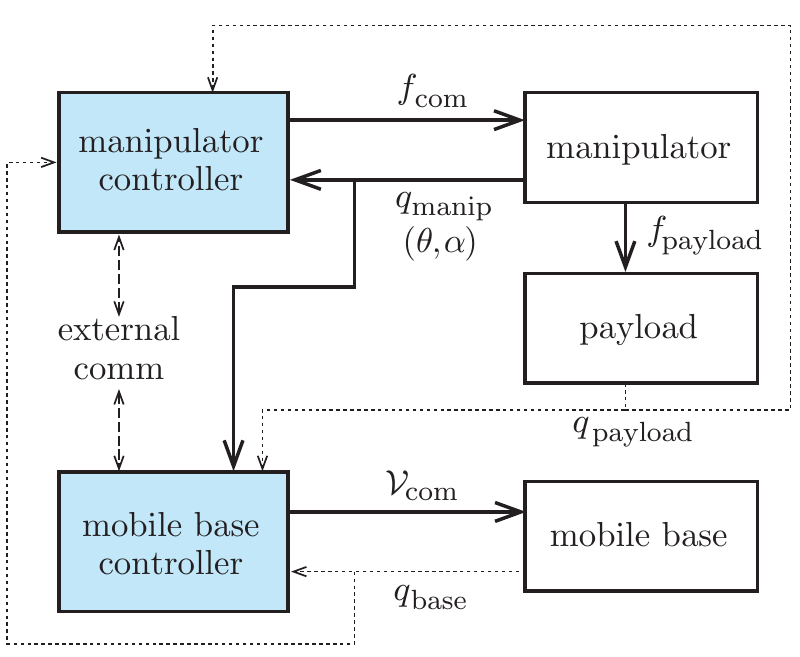}
\end{center}
 \caption{Overview of each Omnid's control system. The solid lines indicate communication channels used in this paper. The dashed lines indicate optional wireless communication with a centralized controller or other Omnids. The dotted lines indicate optional exteroceptive sensing of the payload and mobile base configurations.
 }
 \label{fig:omnid-control}
\end{figure}

The overall control architecture for each Omnid is illustrated in Figure~\ref{fig:omnid-control}. The manipulator controller calculates a commanded end-effector force $f_\mathrm{com}$ as a function of the manipulator's configuration $q_{\textrm{manip}} = (\theta,\alpha)$ relative to the mobile base, and the mobile base controller calculates the commanded chassis-frame planar twist of the mobile base $\twist_{\textrm{com}}$ as a function of $q_{\textrm{manip}}$. With this architecture, we have implemented several useful control behaviors for single Omnids and a collaborating team of Omnids. The two team behaviors that we focus on in this paper are (1) payload float mode and (2) approximate payload float mode, described in detail below. 

\subsection{Payload Float Mode}
\label{ssec:float}

In this mode, the Omnids balance the gravitational force acting on a rigid payload. Thus the human user(s) experience only the inertia of the payload, and are responsible only for the force to accelerate the weightless inertia. The magnitude of this force is typically dwarfed by the gravitational force. 

The manipulator and mobile base controllers that implement payload float mode are described below.

\subsubsection{Manipulator Controller for Payload Float Mode}

In payload float mode, each Omnid's manipulator implements active force control. The commanded end-effector force is
\begin{equation}
    f_{\textrm{com}} = f_{\textrm{manip}} + f_{\textrm{pay}} +f_{\text{rest}} \in \real^3, \label{eq:float}
\end{equation}
where $f_{\textrm{manip}}$ is the force needed to cancel gravity on the manipulator itself, $f_{\textrm{pay}}$ is the additional force needed to cancel a portion of the payload's gravitational force, and $f_{\textrm{rest}}$ is a restoring force that pushes the end-effector toward the workspace center if it approaches a boundary. The SEA joint torques are calculated as $\tau = J^{\intercal}(\theta) f_{\textrm{com}}$, and the Omnids use 800~Hz PID joint controllers to track the commanded torques.

Each of the terms of Equation~\eqref{eq:float} is described below.

\paragraph{Manipulator Gravity Compensation}
The joint-space mass matrix of the manipulator is $\mathbb{M}(\theta,\alpha) \in \real^{3 \times 3}$, which can be expressed in the $(x,y,z)$ task space as $\Lambda(\theta,\alpha) = J^{-\intercal}(\theta) \mathbb{M}(\theta,\alpha) J^{-1}(\theta)$. The end-effector force 
\[
f_{\textrm{manip}} = -\Lambda(\theta,\alpha) 
\begin{bmatrix}
0 \\
0 \\
\mathfrak{g}
\end{bmatrix},
\]
is calculated to cancel the effect of the gravitational acceleration $\mathfrak{g}$ in the $z$ direction. 

\paragraph{Payload Gravity Compensation}
To implement exact payload gravity compensation, each Omnid must know the force component it is responsible for, based on its grasp location relative to the payload's center of mass and the orientation of the payload. In our implementation, the orientation of the payload is sensed by the gimbal wrist and we assume that the grasp location relative to the center of mass is known\footnote{Each Omnid's grasp location relative to the payload's center of mass can be estimated by an initial calibration routine or specified directly.}. On startup, each Omnid measures the force it supports. That information, plus the payload orientation and grasp location, allows each Omnid to solve a statics problem to continually calculate its own $f_{\textrm{pay}}$ throughout manipulation.

\paragraph{Workspace Boundary Repulsion}
Each Delta manipulator has a limited workspace. If the end-effector is within 2~cm of the boundary of a sphere conservatively inscribed in its workspace (Table~\ref{tab:params}), a spring-like restoring force $f_{\textrm{rest}}$ pushes the end-effector toward the center of its workspace. This results in smooth degradation of payload gravity compensation when the manipulators approach their workspace boundaries. 

\subsubsection{Mobile Base Controller for Payload Float Mode}
The mobile base moves to keep the horizontal $(x,y)$ coordinates of the wrist and the gimbal angle $\alpha_z$ near zero, ensuring maximum range of motion of the manipulator to implement force control. This ``recentering control'' uses a PD controller to calculate a twist $\twist_{\textrm{com}} \in \real^3$ to drive $(x,y,\alpha_z)$ to zero.

\subsection{Approximate Payload Float Mode}
\label{ssec:pseudo}
This control mode is used for payloads where exact gravity compensation is not possible because the center of mass is unknown or changing, possibly due to unmodeled internal articulation or flexibility. In this mode, mobile base control is identical to the recentering control of the payload float mode, but the manipulator controller~\eqref{eq:float} is altered to
\begin{equation}
    f_{\textrm{com}} = f_{\textrm{manip}} + f^{'}_{\textrm{pay}} + f_{\textrm{spring}} + f_{\text{rest}} \in \real^3, \label{eq:approxfloat}
\end{equation}
where $f_{\textrm{pay}}$ of Equation~\eqref{eq:float} is replaced by the terms $f^{'}_{\textrm{pay}} + f_{\textrm{spring}}$. These two terms are explained next. 

On startup, each Omnid measures the height of its manipulator $z_0$ and the gravitational load on its manipulator due to the payload $-f^{'}_{\textrm{pay}}$. Throughout the rest of the manipulation, the manipulator provides a constant nominal force $f^{'}_{\textrm{pay}}$, which provides exact payload gravity compensation if the Omnids maintain their relative configuration. Since the Omnids will move, the term $f_{\textrm{spring}} =  c(z_0 -z), c>0$, pulls each Omnid manipulator toward the set height $z_0$, to prevent imperfect gravity compensation from causing the payload to drift to the boundaries of the end-effector workspaces. If $|z_0-z|$ exceeds a threshold $\epsilon$, the set height $z_0$ is changed to maintain $|z_0-z| \leq \epsilon$. This allows the user to drag the payload to a new nominal height at each Omnid, similar to dragging an object over a surface with Coulomb friction by pulling with a spring. 

\subsection{Other Controllers}

Other controllers we have implemented on the Omnids include end-effector impedance control, end-effector motion control, and mobile base trajectory-tracking control. These controllers are demonstrated in the accompanying video~\cite{OmnidVideo}. As these are not used in the human-multimocobot manipulation experiments, we omit their detailed description.

\section{Validation and Demonstration}
\label{sec:experiments}

Please see the accompanying video~\cite{OmnidVideo} for more details.

\subsection{SEA Performance}

\emph{Static torque measurement accuracy}: Static load tests of the SEA indicate that the worst-case error in torque measurement is $\pm 2$\% of the full-scale continuous torque range of $\pm 9$~Nm. 

\emph{Static torque measurement precision}: Any change in torque of $0.01$~Nm ($0.1$\% full scale) or greater is reliably detected.

\begin{figure}
    \centering
    \includegraphics[width=2.7in]{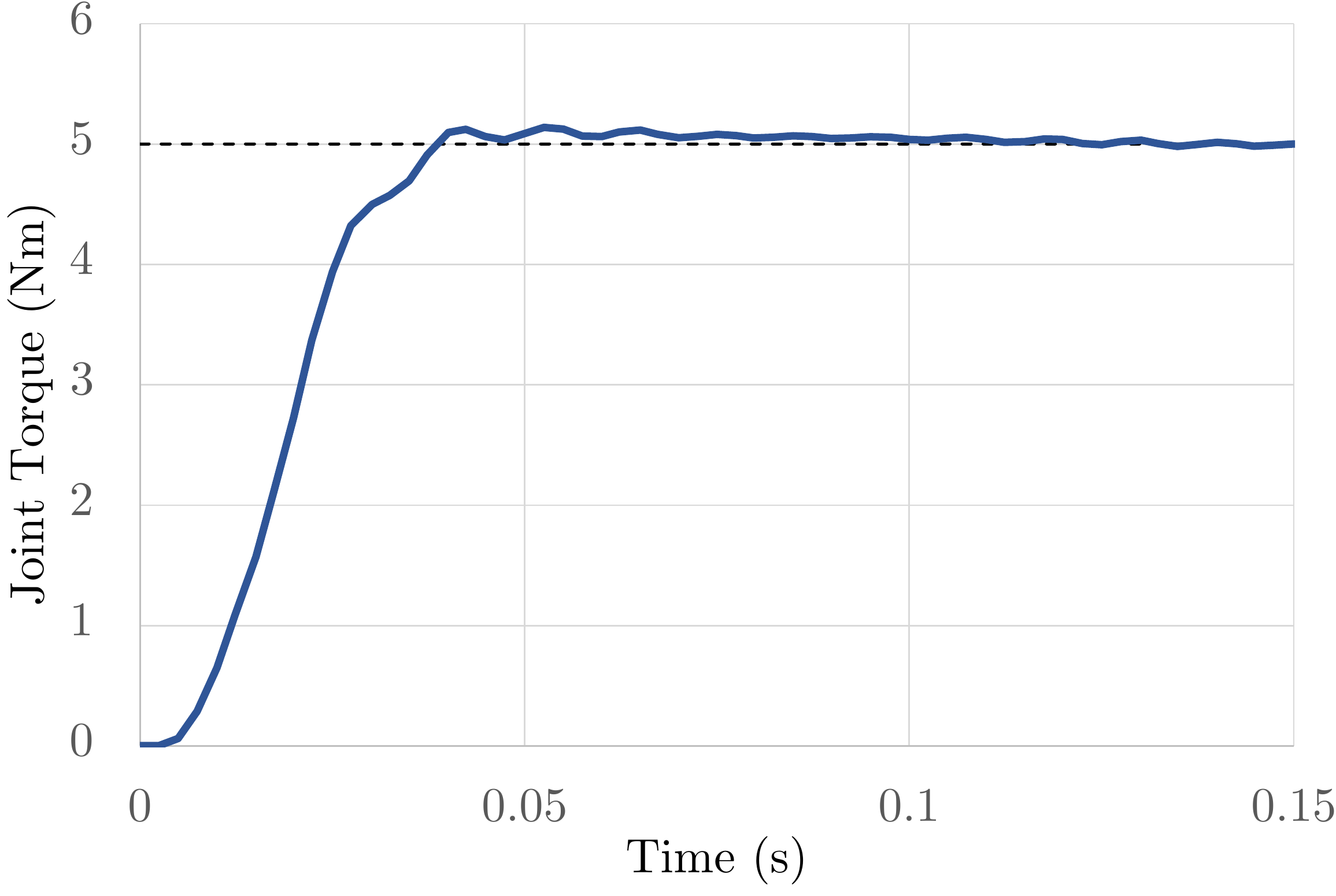}
    \caption{Experimental step response of a blocked SEA joint to a $5$~Nm torque command.}
    \label{fig:step}
\end{figure}

\begin{figure}
    \centering
    \includegraphics[width=2.7in]{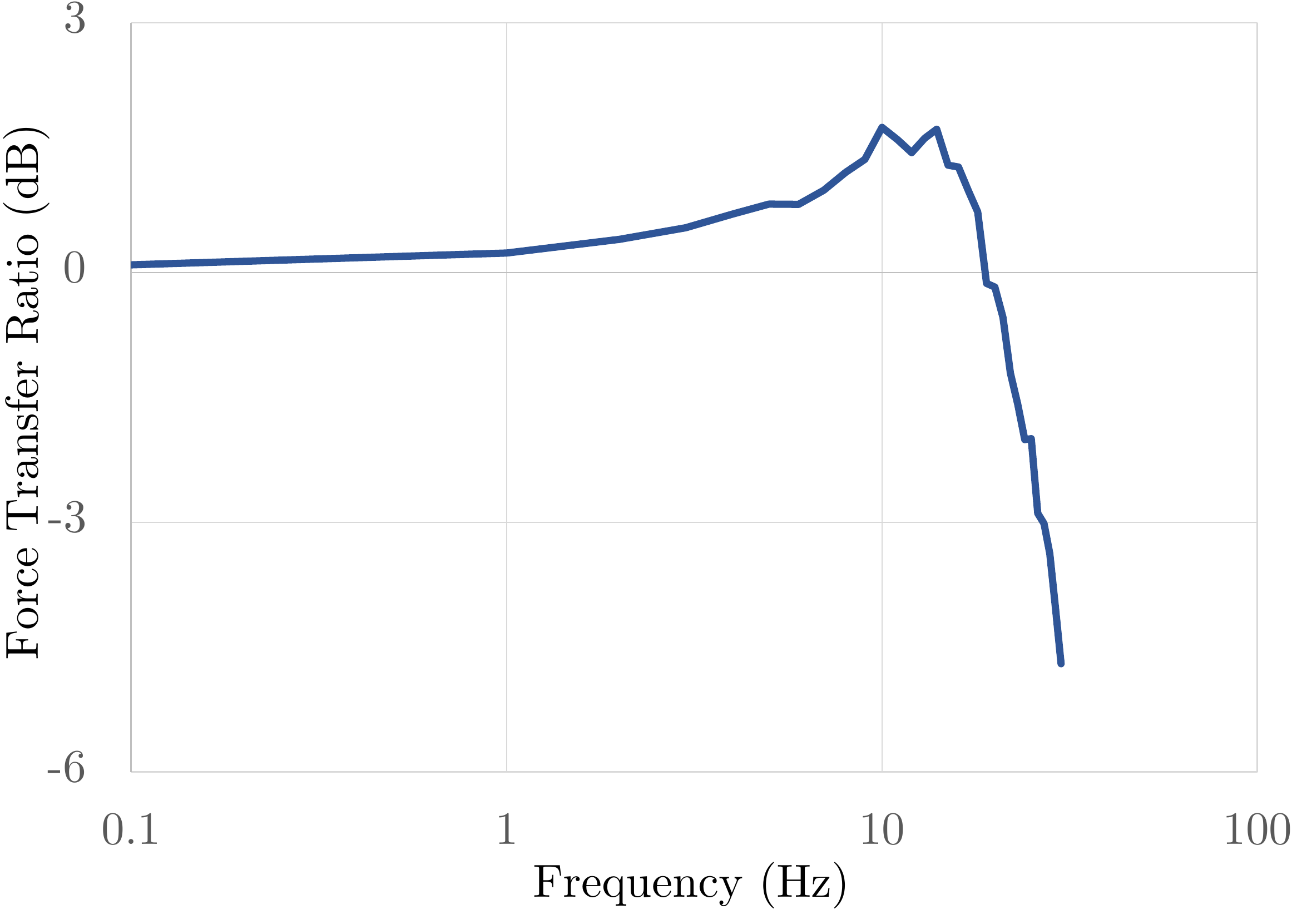}
    \caption{Empirical frequency response of a blocked SEA joint to a $\pm1$~Nm sinusoidal torque command about a DC offset of $5$~Nm.} 
    \label{fig:freq}
%    \vspace{-15pt}
\end{figure}

\emph{SEA torque control step and frequency response}: Figure~\ref{fig:step} shows the measured $5$~Nm step response of a single SEA for a blocked proximal link. The step response settling time is less than $0.1$~s. Figure~\ref{fig:freq} shows the frequency response to oscillating torque commands about a $5$~Nm DC offset. Under these conditions, the bandwidth is between $20$ and $30$~Hz.

\subsection{Single-Omnid Experiments}
Static load tests using weights at the end-effector of an Omnid indicate a worst-case force-sensing error of $\pm 2$\% relative to the manipulator's maximum continuous force of $\pm 90$~N at the home configuration. 

Ideally, when running mobile base recentering control and controlling the end-effector to support its own weight, the end-effector would begin to move in response to any nonzero applied force. In practice, friction causes a deadband where no motion occurs. With the manipulator near its home configuration and a payload of approximately $5$~kg, motion typically occurs at less than $1$~N of applied force.

\begin{figure}
    \centering
    \includegraphics[width=3.2in]{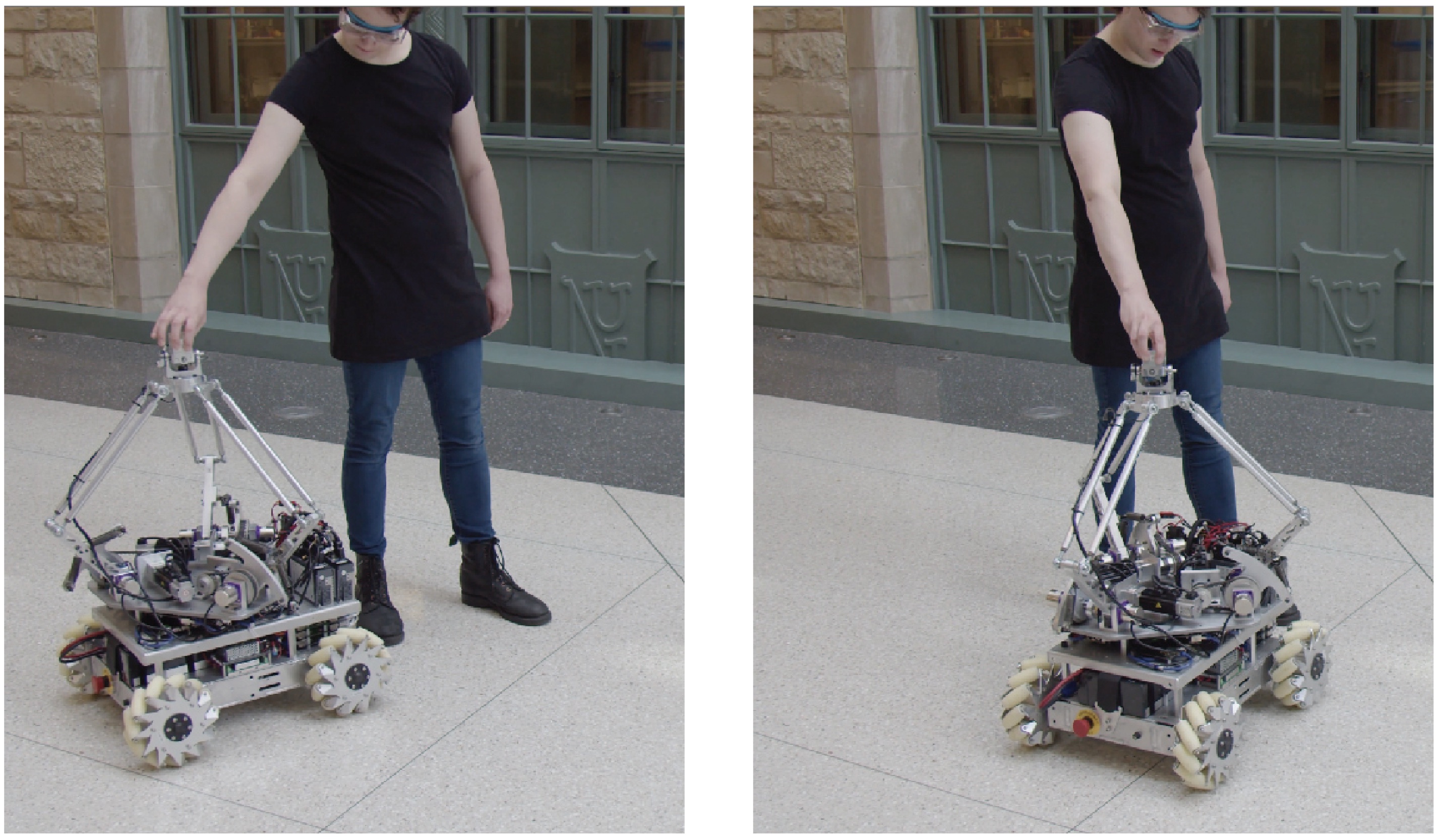}
    \caption{``Walk the dog'': manipulator gravity compensation plus mobile base recentering control allows a human operator to easily move the Omnid with a light touch.}
    \label{fig:walk-dog}
\end{figure}

Figure~\ref{fig:walk-dog} shows the Omnid ``walk the dog'' behavior, where the manipulator implements manipulator gravity compensation and the mobile base implements recentering control. The human repositions the mobile base using only a light touch.

\subsection{Multi-Omnid Single-Human Assembly of a Rigid Payload}
\label{ssec:pvcexp}

In this experiment (Figure~\ref{fig:billie}), a single human and three Omnids collaboratively manipulate a rigid PVC pipe assembly weighing $15.6$~kg, significantly more than a single Omnid's $9$~kg capacity at the manipulator's home configuration. The Omnids run the payload float mode of Section~\ref{ssec:float}, making the pipes feel weightless to the human. In a 6-dof assembly task, where two pipes are inserted into holes with a $2$~mm tolerance, insertion is accomplished easily, dynamically, and intuitively with small operator forces (see~\cite{OmnidVideo}). 

\subsection{Multi-Omnid Multi-Human Manipulation of an Articulated Payload}
\label{ssec:artexp}

\begin{figure}
    \centering
    \includegraphics[width=3.2in]{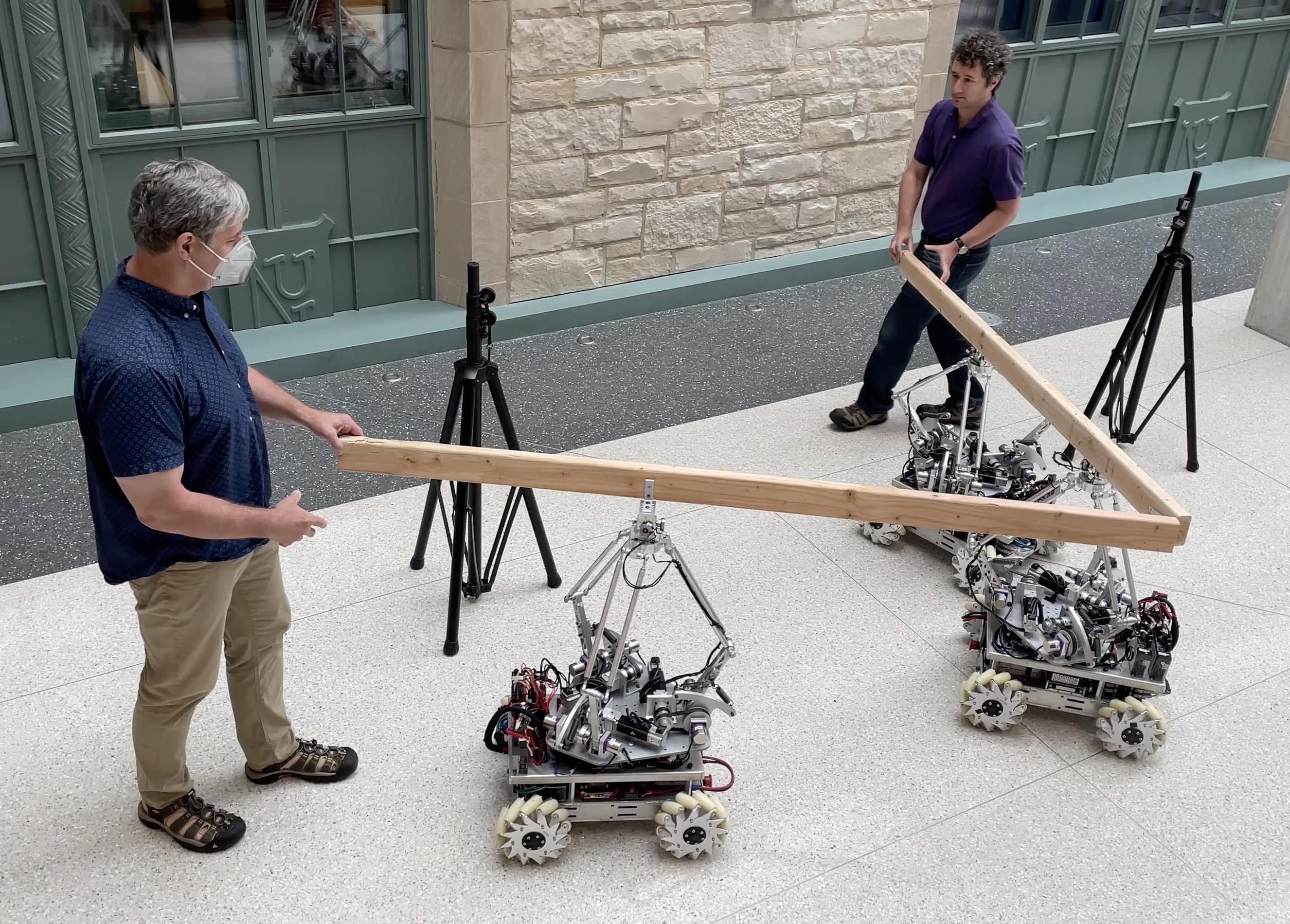}
    \caption{Two humans and three Omnids manipulating a jointed payload.}
    \label{fig:articulated}
%    \vspace{-15pt}
\end{figure}

Figure~\ref{fig:articulated} shows two humans and three Omnids collaboratively manipulating an articulated payload among obstacles. The payload has seven total degrees of freedom: six for one rigid body and one for the hinge joint. Three Omnids are sufficient to autonomously manipulate this payload (Table~\ref{table:manipulability}), but a singularity occurs if the three wrists are brought into alignment: the Omnids cannot generate a torque about the line, so the human operators must provide it.

In this task, the Omnids implement approximate payload float mode (Section~\ref{ssec:pseudo}), since the payload's center of mass is unknown and changing due to the payload's unmodeled articulation. Our experiments show that one or two human operators can easily and dynamically control all seven degrees of freedom of the articulated payload (see~\cite{OmnidVideo}).

\subsection{Discussion: Manipulation of a Weightless Payload}
In our proof-of-concept demonstrations of collaborative manipulation in payload float mode (Sections~\ref{ssec:pvcexp} and \ref{ssec:artexp}), we found that new users required no training to manipulate large payloads effortlessly and fluidly. Because the interaction is direct, physical, and familiar, no new mental mappings need to be learned, resulting in insignificant cognitive loading relative to most forms of teleoperation (e.g.,~\cite{triantafyllidis2020study}). The gravity cancellation provided by the Omnids simply allows users to apply drastically smaller forces to manipulate the payload. A formal study quantifying intuitiveness of physical co-manipulation versus remote control left for future work.

\section{Conclusion and Future Work}
In this paper, we introduced the concept of a mocobot team and its design requirements, including passive compliance, payload manipulability, and manipulation force control decoupled from mobile base motion control. We also introduced the design and control of a team of Omnid mocobots which meet the design requirements, and {initial experiments indicate} that their payload float modes allow one or more humans to easily, dynamically, and intuitively manipulate large, awkward, and articulated payloads. To the best of our knowledge, the Omnids are the first mocobot team to implement significant passive compliance and three-dimensional force-controlled weightlessness for human-multirobot collaborative mobile manipulation.

The Omnids are a platform for evaluating controllers for human-multirobot collaborative manipulation, not an industry-ready solution. For example, the Omnid manipulators have a limited workspace and cannot autonomously grasp a payload; the payload is manually attached to the robots before each experiment. The control strategies described here, however, may apply to other mocobot designs, including outdoor and legged mocobots, provided they satisfy the design criteria described in Section~\ref{sec:considerations}.

Future work includes: (1) new mocobot designs with larger manipulator workspaces and mobile bases suitable for outdoor use; (2) performance studies of transport and assembly tasks by first-time and experienced humans under different control schemes (e.g., teleoperation and direct human-payload physical interaction), to further explore the benefits of haptic interaction; (3) advanced collaboration capabilities enabled by wireless communication between the mocobots; and
(4) autonomous multirobot manipulation, including estimation of payload kinematic and inertial properties and payload-centric wrench and impedance control for assembly tasks. 

\bibliographystyle{IEEEtran}
\bibliography{main.bib}

\end{document}